\def\year{2019}\relax
\documentclass[letterpaper]{article} 
\usepackage{aaai19}  
\usepackage{times}  
\usepackage{helvet}  
\usepackage{courier}  
\usepackage{url}  
\usepackage{graphicx}  
\usepackage{todonotes}
\usepackage{amsmath}
\usepackage{amssymb}
\usepackage{amsthm}
\usepackage{mathtools}
\usepackage{tabularx}
\usepackage{caption}
\begin{document}
%
\title{Discourse Analysis for Evaluating Coherence in Video Paragraph Captions}
\author{ Arjun R. Akula, Song-Chun Zhu \\
UCLA Center for Vision, Cognition, Learning, and Autonomy\\
aakula@ucla.edu, sczhu@stat.ucla.edu 
}
\maketitle
\begin{abstract}
Video paragraph captioning is the task of automatically generating a coherent paragraph description of the actions in a video. Previous linguistic studies have demonstrated that coherence of a natural language text is reflected by its discourse structure and relations. However, existing video captioning methods evaluate the coherence of generated paragraphs by comparing them merely against human paragraph annotations and fail to reason about the underlying discourse structure. At UCLA, we are currently exploring a novel discourse based framework to evaluate the coherence of video paragraphs. Central to our approach is the discourse representation of videos, which helps in modeling coherence of paragraphs conditioned on coherence of videos. We also introduce DisNet, a novel dataset containing the proposed visual discourse annotations of 3000 videos and their paragraphs. Our experiment results have shown that the proposed framework evaluates coherence of video paragraphs significantly better than all the baseline methods. We believe that many other multi-discipline Artificial Intelligence problems such as Visual Dialog and Visual Storytelling would also greatly benefit from the proposed visual discourse framework and the DisNet dataset.
\end{abstract}

\section{Introduction}
In recent years, video captioning task has drawn a lot of attention from computer vision community where the task is to explain video semantics using sentence descriptions (video-paragraphs)~\cite{guadarrama2013youtube2text,thomason2014integrating,yu2016video}. With the introduction of large scale video captioning datasets, great advances have been achieved in describing the videos using multiple-sentence descriptions. While the success of these methods is encouraging, they all lack a formal approach to compute coherence of video-paragraphs~\cite{DBLP:journals/corr/abs-1903-02252,carlson2003building,soricut2003sentence,lethanh2004generating}. 

At UCLA, we are working on a novel discourse based framework to model the coherence in video-captioning task. Text-level discourse structure aids in understanding a piece of text by linking it with other text units (such as surrounding clauses, sentences, etc)~\cite{carlson2003building,soricut2003sentence,lethanh2004generating}. Linguistic studies demonstrated that coherence of a text is reflected by its discourse structure and relations.  In other words, coherent text favors certain types of discourse relation transitions.

\begin{figure}[t]
\centering
  \includegraphics[width=\linewidth]{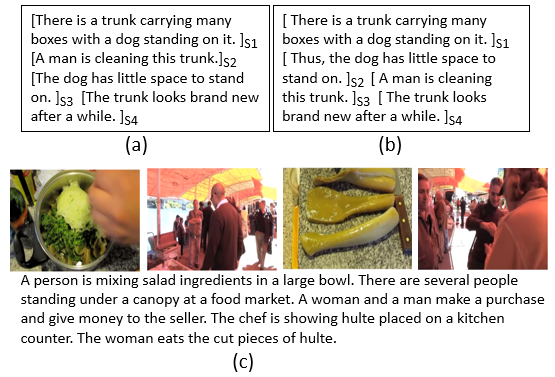}
  \caption{(a) Incoherent video-paragraph; (b) Coherent video-paragraph; (c) Incoherent video-paragraph but becomes coherent when conditioned on the coherence of input video.}~\label{fig:intro1}
\end{figure}

For example, video-paragraph shown in Figure~\ref{fig:intro1}(b) is clearly more coherent than Figure~\ref{fig:intro1}(a). This can be attributed to the ordering of the discourse relations among the sentences. In Figure~\ref{fig:intro1}(a), the sentence S1 is the cause for sentence S3; S2 is the cause for S4. In Figure~\ref{fig:intro1}(a), the sentence S1 is the cause for sentence S2; S3 is the cause for S4. The discourse relation `cause' is more localized in Figure~\ref{fig:intro1}(b) and is therefore more coherent compared to Figure~\ref{fig:intro1}(a). Existing video captioning methods fail to perform such discourse based analysis to assess the coherence of generated video-paragraphs. Instead they rely on metrics such as BLEU, METEOR, ROUGE, CIDEr, etc which merely compare the generated video-paragraphs with human paragraph annotations~\cite{DBLP:journals/corr/abs-1903-02252,akula20words,akula2019visual,akula2021crossvqa,akula2021measuring,akula2021robust,akula2020cocox,r2019natural,pulijala2013web,gupta2012novel}.

\textbf{Can human-generated annotations be incoherent?} We did an experiment to verify whether if it is possible for humans to write incoherent descriptions. This is an important question because, if the human paragraphs are always coherent, then we can infer coherency of generated paragraphs by simply using scores such as BLEU. However, if human paragraphs are not always coherent, then relying only on scores such as BLEU is problematic and hence requires explicit coherence assessment methods like ours. We conducted a human study experiments where we asked people to describe 100 videos (randomly chosen from ActivityNet dataset~\cite{}) without training them on how to write coherent paragraphs. We then evaluated the coherency of the descriptions using different set of human subjects. We found that about 72\% of the descriptions are found to be incoherent~\cite{akula2013novel,akula2018analyzing,akula2021mind,gupta2016desire,akula2019explainable,akula2021gaining,akula2019x,akula2020words}.

Moreover, video-paragraphs may sometimes appear to be incoherent linguistically but are actually coherent given the video content. For example, video-paragraph shown in Figure~\ref{fig:intro1}(c) looks incoherent as the topics across the sentences shift abruptly. However, note that, this is one of the best possible coherent description one can generate for the video shown in Figure~\ref{fig:intro1}(c). This implies that video-paragraph, in this case, is indeed coherent whereas the underlying video is incoherent. In this work, we identify such cases by proposing a discourse representation for videos. We model the coherence of the video-paragraphs by conditioning on coherence of their corresponding videos. We also introduce DisNet, a novel dataset containing discourse annotations of about 3000 videos, to learn a model to automatically infer visual discourse relations. To the best of our knowledge, this is the first work to introduce discourse framework for videos~\cite{akula2022effective,agarwal2018structured,akula2019natural,akula2015novel,palakurthi2015classification,agarwal2017automatic,dasgupta2014towards}.

\textbf{Contributions.} The contributions of this work are threefold: (i) novel discourse representation of videos and video-paragraphs; (ii) a new dataset containing discourse annotations of videos and video-paragraphs; (iii) a novel approach to assess the coherence of video-paragraphs based on discourse cues from both videos as well as video-paragraphs.

The remainder of this paper is organized as follows: First, we define the discourse representation of videos and video-paragraphs in video-captioning task. Next, we introduce a new dataset containing the proposed discourse annotations. We also train models using this dataset and report the baseline accuracies in automatically inferring the discourse representations of videos and video-paragraphs. Finally, in the experiments section, we demonstrate that the proposed discourse framework is effective in evaluating coherence of paragraphs generated from video-captioning task.\\

\section{Visual Discourse Representation} 
\textbf{Videos.} Defining discourse representation of videos comprises three main steps, namely: (a) identification of elementary discourse units (EDUs)~\cite{mann1988rhetorical}; (b) identification of discourse acts, which are categories of EDUs that pertain to their role in the videos and (c) identification of discourse relations, which aid in understanding relationships among EDUs. In our pilot experiments, we found that shots are effective EDUs for videos. A shot in a video refers to a contiguous recording of one or more frames depicting a continuous action in space and time. Therefore, we first segment the video into shots and then define the discourse acts and relations among shots. We used the dense captioning events dataset to sample videos and classified shots in the videos using a manual iterative process with experts until a stable set of discourse acts and relations are obtained.

We define the discourse structure of video as directed acyclic graph $\mathbf{G}$ consisting of a set of nodes $\mathbf{V}$ and a set of labelled edges $\mathbf{E}$. The interior nodes (non-terminal nodes) denote the discourse acts and the leaf nodes (terminal nodes) constitute the shots obtained from video segmentation. $\mathbf{P}$ denotes the root node that doesn't have any branches on top of it. Each video contains only one root node. The edges connecting interior nodes and leaf nodes denote the discourse relations. Detailed information about each discourse act and the relations
allowed are given below.\\
\textit{Discourse Acts:} We found the following three discourse acts to be helpful in analyzing the coherence of the video:\\
(a) Primary Context: Shots in the video that contain the main theme/information are labelled as primary context.\\
(b) Secondary Context: Shots in the videos that aid in understanding the primary context of the video are labelled as secondary context.\\ 
(c) Auxiliary Context: Shots in the videos that are not essential in interpreting the primary context of the video are labelled as auxiliary context. \\
For instance, consider a typical sports video which usually contain shots related to players, scores and crowd. Shots showing the players - which is the main theme in this case - is labelled as primary context; shots showing scores are labelled as secondary context and shots showing crowd are labelled as auxiliary context. In our experiments, we found that videos that have only primary context are usually more coherent than videos that have all the three (primary, secondary, auxiliary) contexts.\\\\
\textit{Discourse Relations:} In addition to the three discourse acts, we found the following six discourse relationships to be crucial in analyzing the coherence of videos:\\
(a) Interpretation: The edge label between a shot and a discourse act is labelled as interpretation if the shot summarizes or helps in interpreting/establishing the context of the discourse act. For a given discourse act in the video discourse structure, there can be only one interpretation. Therefore, for most of the videos, the interpretation is usually (but not necessarily) the shot that appears first in the video among all the shots in that discourse act.\\ 
(b) Sequence: The edge label between a shot and a discourse act is labelled as sequence if the shot introduces new objects that are not yet introduced in the previous shots of discourse act.\\
(c) Sub-Context: The edge label between a shot and a discourse act is labelled as sub-context if the shot gives detailed view of objects that are already introduced briefly in the previous shots of discourse act. For instance, consider a interpretation shot showing a person playing the guitar. If the next shot zooms in to the guitar to give more details of playing guitar, we label this new shot using sub-context relationship.\\
(d) Super-Context: Super-context is a special case of sequence relationship. A shot in super-context relationship introduces new objects and do not shift the focus from objects shown in the previous shot. For instance, zooming out gives the super-context of a shot.\\
(e) Elaboration: The edge label between a shot and a discourse act is labelled as elaboration if the shot continues to focus on objects that are already introduced in the previous shots of discourse act.\\
(f) Repetition: The edge label between a shot and a discourse act is labelled as repetition if a shot is repeated from the previously introduced shots of a discourse act. \\ \\
\textbf{Video-Paragraphs.} We found that only a subset of discourse relations (cause, background, sequence, interpretation and elaboration) from a text-level discourse framework called Rhetorical Structure Theory (RST) are sufficient enough to analyze coherence of paragraphs in video captioning task. Moreover, we found the following five additional new discourse relations among sentences in video-paragraphs, which are specific to video-captioning task and hence are not part of RST framework:\\
(a) Expectation: Discourse relation `Expectation' captures the scenarios where an object/action that is not present in the video but is mentioned in the video-paragraph. For example, in the below video-paragraph, S2 is related to S1 using an Expectation discourse relationship.\\ 
\big[It looks like an office.\big]$_{S1}$ \big[There is no chair and desk.\big]$_{S2}$\\
(b) Sentiment: Sometimes sentences in the video-paragraph mention the sentiment of the video. For instance, in the following video-paragraph, S2 is related to S1 with a sentiment discourse relationship.\\
\big[There are two persons standing next to a bus.\big]$_{S1}$ \big[The scene looks sad.\big]$_{S2}$\\
(c) Continuation: If the sentence continues to describe the same object or event as in the previous sentence, we use continuation relationship to connect them.\\
(d) Parallel: In a video-paragraph, if two sentences describe different activities happening simultaneously in the video, we connect the sentences using parallel discourse relationship.\\
(e) Video-Attribute: It is common in video captioning task to describe the properties of the video in the video-paragraph. For instance, in the below video-paragraph, S2 is related to S1 with a video-attribute relationship.\\
\big[There are two persons standing next to a bus.\big]$_{S1}$ \big[The scene looks blurry.\big]$_{S2}$\\

\section{DisNet Dataset}
We use the 10,000 videos from the newly-released ActivityNet Captions dataset and annotate their discourse structures. ActivityNet Captions contains a total of 20k videos amounting to 800 video hours with 100k total descriptions. These videos are diverse and comprehensive and their video-paragraphs are collected by training the human subjects to write coherent descriptions. Therefore their video-paragraphs are mostly coherent. Therefore they are well-suited for our discourse representation task. Annotation of discourse acts and relations was conducted by eight graduate and undergraduate students. Each video and its video-paragraph was annotated by at least two students. We solved the disagreements found in the annotations together. Before students began annotating, they were presented with an instruction manual that explained definitions of discourse acts and relations. They were also given a few warm-up videos and video-paragraphs to do as practice before beginning annotation.

As our goal is to better understand the discourse structure, we chose to only take videos that have at least 3 shots. Also, some videos have hundreds of shots. As this is too much work for an annotator, we discarded the videos with more than 30 shots. During annotation, few more videos were discarded due to bugs that occurred with the shot detector algorithm. Finally, at the end, 3000 videos and the corresponding 3000 video-paragraphs were fully annotated with their discourse structures. We will now present an overview of the dataset.

\subsection{Dataset Statistics}
On average, each of the 3000 videos have about 12.6 shots. We found that, as the number of discourse acts in the video increases, the number of shots also increases.  The average duration of the videos is 120 seconds. The average number of sentences in the 3000 video-paragraphs is 3.12. We found that 67\% of the shots in the dataset are annotated as primary context, 14\% as secondary context and 19\% as auxiliary context. The inter-rater reliability (using Krippendorf's Alpha) is 0.81 for primary context, 0.62 for secondary context and 0.78 for auxiliary context. Furthermore, videos containing all the three discourse acts, usually but not always, are found to be less coherent than videos containing only the primary context. 

\begin{table}[t]
    \caption {Percentage and inter-rater reliability of discourse relations in video-paragraphs}
    
    \begin{minipage}{.8\linewidth}
      \centering
        
\begin{tabular}{|m{1.75cm}|m{0.55cm}|m{0.5cm}|}
\hline
Discourse \break Relation & Total \% & $\alpha$\\ 
\hline
Background & 4.3 & 0.66\\
\hline
Continuation & 12.7 & 0.57\\
\hline
Parallel & 16.3 & 0.66\\
\hline
Elaboration & 8.0 & 0.62\\
\hline
Cause & 7.1 & 0.38\\
\hline
Video-Attribute & 4.1 & 0.89\\
\hline
Sequence & 25.4 & 0.81\\
\hline
Interpretation & 15.8 & 0.74\\
\hline
Sentiment & 2.8 & 0.72\\
\hline
Expectation & 3.5 & 0.62\\
\hline
\end{tabular}
\label{tab1}
    \end{minipage} 
\end{table}

Table~\ref{tab1} shows the proportion of video discourse relations in our dataset. As can be seen, the relations elaboration, sequence and interpretation make up a large portion of the dataset. The inter-rater reliability show that the relation annotations for videos were almost entirely in agreement. The relations sequence, interpretation, parallel and continuation dominate other relations in the dataset. In particular, the interpretation relation is found at least once in most paragraphs. This can be attributed to the fact that most paragraphs in the dataset begin with a brief summary of video. The inter-rater reliability of these relations show that some relations had more agreement between annotators than others. The relation `cause' was the least reliable category. This was perhaps due to the lack of explicit connectives (such as because) in the paragraph descriptions. We also noticed that elaboration relation had some overlap with sequence and background relations.



\subsection{Predicting Discourse Acts and Relations in Videos}
We use standard machine translation encoder-decoder RNN model (Sutskever et al., 2014) for learning to predict the sequence of discourse acts from the input sequence of shots in videos. As RNN suffers from decaying of gradient and blowing-up of gradient problem, we use LSTM units, which are good at memorizing long-range dependencies due to forget-style gates (Hochreiter and Schmidhuber, 1997). The sequence of video shots are passed to the encoder. The last hidden state of the encoder is then passed to the decoder and then the decoder generates the sequence of discourse acts. Let the input sequence of video shots be $\mathbf{x} = \left(x_1,...,x_p\right)$ and the output sequence of discourse acts as $\mathbf{y} = \left(y_1,...,y_n\right)$. The distribution of the output sequence w.r.t. the input sequence is:

\begin{equation}
p(y_1,...,y_n \mid x_1,...,x_p) = \prod_{t=1}^{n} p(y_t\mid h_t^d)
\end{equation}
where $h_t^d$ is the hidden state at the $t^{th}$ time step of the decoding LSTM.
We further improve our encoder-decoder model using an attention based sequence-to-sequence model~\cite{bahdanau2014neural}. The attention weights act as an alignment mechanism by re-weighting the encoder hidden states that are more relevant for decoder time step.

We learn discourse relations in videos using an other encoder-decoder RNN model similar to the discourse act prediction model. However, in this case, we first predict the discourse acts and then use them as input features in addition to the input video scenes.

We use 210 videos for training, 30 videos for validation and the rest 70 videos for testing. We fix our sampling rate to 5fps to bring uniformity in the temporal representation of actions across all videos. These sampled frames are then converted into features using VGGNet~\cite{simonyan2014very}. We initialize the decoder embedding with Google pre-trained word2vec word embeddings~\cite{mikolov2013distributed}. We tune all hyperparameters using our validation data: learning rate, weight initializations, hidden states. We use a 1024-dimension RNN hidden state size. We use Adam optimizer~\cite{kingma2014adam}. We apply a dropout of 0.5.

\textbf{Results}:Attention based models gave better accuracies than the simple sequence to sequence models. In particular, the LSTM unit with 512 hidden units and single encoding layer outperformed other configurations. These results are encouraging considering the small size of our training dataset.

\subsection{Predicting Discourse Relations in Video-Paragraphs}
We use RST based disocourse representation to model discourse relations of video paragraphs as many of our relations belong to RST framework. We use shift-reduce parser to represent the discourse relations in RST framework. Shift-reduce discourse parser is an incremental discourse parsing approach,
that maintains a queue containing the EDUs in the paragraph that have not been processed yet, and a stack of RST subtrees that will eventually be combined to form a complete tree. All EDUs are placed in the queue and the stack is empty initially. The parser selects shift or reduce actions in an iterative manner until a complete RST tree is found or no further actions can be executed. The problem of selecting the best parsing action given the current parsing state (i.e.,
the stack and queue) is modeled as a max margin classification problem.



We use 2200 video paragraphs for training, 200 for validations and the rest 600 for testing. We evaluate the performance on test data using F1 scores for agreement on unlabeled EDU spans, spans labeled only with nuclearity, and fully
labeled spans (i.e. including relation information). The F1-score is 84.6\% for unlabeled spans, 71.6\% for labeled spans and 45.9\% for fully labeled spans.







\section{Conclusion}
We are currently exploring a novel discourse based framework to evaluate the coherence of video paragraphs. Central to our approach is the discourse representation of videos, which helps in modeling coherence of paragraphs conditioned on coherence of videos. We also introduce DisNet, a novel dataset containing the proposed visual discourse annotations of 3000 videos and their paragraphs.
\bibliography{main.bib}

\begin{thebibliography}{}

\bibitem[\protect\citeauthoryear{Agarwal \bgroup et al\mbox.\egroup
  }{2017}]{agarwal2017automatic}
Agarwal, S.; Aggarwal, V.; Akula, A.~R.; Dasgupta, G.~B.; and Sridhara, G.
\newblock 2017.
\newblock Automatic problem extraction and analysis from unstructured text in
  it tickets.
\newblock {\em IBM Journal of Research and Development} 61(1):4--41.

\bibitem[\protect\citeauthoryear{Agarwal \bgroup et al\mbox.\egroup
  }{2018}]{agarwal2018structured}
Agarwal, S.; Akula, A.~R.; Dasgupta, G.~B.; Nadgowda, S.~J.; and Nayak, T.~K.
\newblock 2018.
\newblock Structured representation and classification of noisy and
  unstructured tickets in service delivery.
\newblock US Patent 10,095,779.

\bibitem[\protect\citeauthoryear{Akula and
  Zhu}{2019a}]{DBLP:journals/corr/abs-1903-02252}
Akula, A.~R., and Zhu, S.
\newblock 2019a.
\newblock Visual discourse parsing.
\newblock {\em CVPR 2019 Workshop on Language and Vision, arXiv:1903.02252}.

\bibitem[\protect\citeauthoryear{Akula and Zhu}{2019b}]{akula2019visual}
Akula, A.~R., and Zhu, S.-C.
\newblock 2019b.
\newblock Visual discourse parsing.
\newblock {\em ArXiv preprint} abs/1903.02252.

\bibitem[\protect\citeauthoryear{Akula and Zhu}{2022}]{akula2022effective}
Akula, A., and Zhu, S.-C.
\newblock 2022.
\newblock Effective representation to capture collaboration behaviors between
  explainer and user.
\newblock {\em arXiv preprint arXiv:2201.03147}.

\bibitem[\protect\citeauthoryear{Akula \bgroup et al\mbox.\egroup
  }{2019a}]{akula2019x}
Akula, A.~R.; Liu, C.; Saba-Sadiya, S.; Lu, H.; Todorovic, S.; Chai, J.~Y.; and
  Zhu, S.-C.
\newblock 2019a.
\newblock X-tom: Explaining with theory-of-mind for gaining justified human
  trust.
\newblock {\em arXiv preprint arXiv:1909.06907}.

\bibitem[\protect\citeauthoryear{Akula \bgroup et al\mbox.\egroup
  }{2019b}]{akula2019explainable}
Akula, A.~R.; Liu, C.; Todorovic, S.; Chai, J.~Y.; and Zhu, S.-C.
\newblock 2019b.
\newblock Explainable ai as collaborative task solving.
\newblock In {\em CVPR Workshops},  91--94.

\bibitem[\protect\citeauthoryear{Akula \bgroup et al\mbox.\egroup
  }{2019c}]{akula2019natural}
Akula, A.~R.; Todorovic, S.; Chai, J.~Y.; and Zhu, S.-C.
\newblock 2019c.
\newblock Natural language interaction with explainable ai models.
\newblock In {\em CVPR Workshops},  87--90.

\bibitem[\protect\citeauthoryear{Akula \bgroup et al\mbox.\egroup
  }{2020a}]{akula20words}
Akula, A.~R.; Gella, S.; Al-Onaizan, Y.; Zhu, S.-C.; and Reddy, S.
\newblock 2020a.
\newblock Words aren't enough, their order matters: On the robustness of
  grounding visual referring expressions.
\newblock In {\em ACL}.

\bibitem[\protect\citeauthoryear{Akula \bgroup et al\mbox.\egroup
  }{2020b}]{akula2020words}
Akula, A.~R.; Gella, S.; Al-Onaizan, Y.; Zhu, S.-C.; and Reddy, S.
\newblock 2020b.
\newblock Words aren't enough, their order matters: On the robustness of
  grounding visual referring expressions.
\newblock {\em arXiv preprint arXiv:2005.01655}.

\bibitem[\protect\citeauthoryear{Akula \bgroup et al\mbox.\egroup
  }{2021a}]{akula2021mind}
Akula, A.; Gella, S.; Wang, K.; Zhu, S.-c.; and Reddy, S.
\newblock 2021a.
\newblock Mind the context: The impact of contextualization in neural module
  networks for grounding visual referring expressions.
\newblock In {\em Proceedings of the 2021 Conference on Empirical Methods in
  Natural Language Processing},  6398--6416.

\bibitem[\protect\citeauthoryear{Akula \bgroup et al\mbox.\egroup
  }{2021b}]{akula2021robust}
Akula, A.; Jampani, V.; Changpinyo, S.; and Zhu, S.-C.
\newblock 2021b.
\newblock Robust visual reasoning via language guided neural module networks.
\newblock {\em Advances in Neural Information Processing Systems} 34.

\bibitem[\protect\citeauthoryear{Akula \bgroup et al\mbox.\egroup
  }{2021c}]{akula2021crossvqa}
Akula, A.~R.; Changpinyo, B.; Gong, B.; Sharma, P.; Zhu, S.-C.; and Soricut, R.
\newblock 2021c.
\newblock Crossvqa: Scalably generating benchmarks for systematically testing
  vqa generalization.

\bibitem[\protect\citeauthoryear{Akula \bgroup et al\mbox.\egroup
  }{2021d}]{akula2021measuring}
Akula, A.~R.; Dasgupta, G.~B.; Ekambaram, V.; and Narayanam, R.
\newblock 2021d.
\newblock Measuring effective utilization of a service practitioner for ticket
  resolution via a wearable device.
\newblock US Patent 10,929,264.

\bibitem[\protect\citeauthoryear{Akula, Dasgupta, and
  Nayak}{2018}]{akula2018analyzing}
Akula, A.~R.; Dasgupta, G.~B.; and Nayak, T.~K.
\newblock 2018.
\newblock Analyzing tickets using discourse cues in communication logs.
\newblock US Patent 10,067,983.

\bibitem[\protect\citeauthoryear{Akula, Sangal, and
  Mamidi}{2013}]{akula2013novel}
Akula, A.; Sangal, R.; and Mamidi, R.
\newblock 2013.
\newblock A novel approach towards incorporating context processing
  capabilities in nlidb system.
\newblock In {\em Proceedings of the sixth international joint conference on
  natural language processing},  1216--1222.

\bibitem[\protect\citeauthoryear{Akula, Wang, and Zhu}{2020}]{akula2020cocox}
Akula, A.~R.; Wang, S.; and Zhu, S.
\newblock 2020.
\newblock Cocox: Generating conceptual and counterfactual explanations via
  fault-lines.
\newblock In {\em The Thirty-Fourth {AAAI} Conference on Artificial
  Intelligence, {AAAI} 2020, The Thirty-Second Innovative Applications of
  Artificial Intelligence Conference, {IAAI} 2020, The Tenth {AAAI} Symposium
  on Educational Advances in Artificial Intelligence, {EAAI} 2020, New York,
  NY, USA, February 7-12, 2020},  2594--2601.
\newblock {AAAI} Press.

\bibitem[\protect\citeauthoryear{Akula}{2015}]{akula2015novel}
Akula, A.~R.
\newblock 2015.
\newblock A novel approach towards building a generic, portable and contextual
  nlidb system.
\newblock {\em International Institute of Information Technology Hyderabad}.

\bibitem[\protect\citeauthoryear{Akula}{2021}]{akula2021gaining}
Akula, A.~R.
\newblock 2021.
\newblock {\em Gaining Justified Human Trust by Improving Explainability in
  Vision and Language Reasoning Models}.
\newblock Ph.D. Dissertation, UCLA.

\bibitem[\protect\citeauthoryear{Bahdanau, Cho, and
  Bengio}{2014}]{bahdanau2014neural}
Bahdanau, D.; Cho, K.; and Bengio, Y.
\newblock 2014.
\newblock Neural machine translation by jointly learning to align and
  translate.
\newblock {\em arXiv preprint arXiv:1409.0473}.

\bibitem[\protect\citeauthoryear{Carlson, Marcu, and
  Okurowski}{2003}]{carlson2003building}
Carlson, L.; Marcu, D.; and Okurowski, M.~E.
\newblock 2003.
\newblock Building a discourse-tagged corpus in the framework of rhetorical
  structure theory.
\newblock In {\em Current and new directions in discourse and dialogue}.
  Springer.
\newblock  85--112.

\bibitem[\protect\citeauthoryear{Dasgupta \bgroup et al\mbox.\egroup
  }{2014}]{dasgupta2014towards}
Dasgupta, G.~B.; Nayak, T.~K.; Akula, A.~R.; Agarwal, S.; and Nadgowda, S.~J.
\newblock 2014.
\newblock Towards auto-remediation in services delivery: Context-based
  classification of noisy and unstructured tickets.
\newblock In {\em International Conference on Service-Oriented Computing},
  478--485.
\newblock Springer.

\bibitem[\protect\citeauthoryear{Guadarrama \bgroup et al\mbox.\egroup
  }{2013}]{guadarrama2013youtube2text}
Guadarrama, S.; Krishnamoorthy, N.; Malkarnenkar, G.; Venugopalan, S.; Mooney,
  R.; Darrell, T.; and Saenko, K.
\newblock 2013.
\newblock Youtube2text: Recognizing and describing arbitrary activities using
  semantic hierarchies and zero-shot recognition.
\newblock In {\em Proceedings of the IEEE international conference on computer
  vision},  2712--2719.

\bibitem[\protect\citeauthoryear{Gupta \bgroup et al\mbox.\egroup
  }{2012}]{gupta2012novel}
Gupta, A.; Akula, A.; Malladi, D.; Kukkadapu, P.; Ainavolu, V.; and Sangal, R.
\newblock 2012.
\newblock A novel approach towards building a portable nlidb system using the
  computational paninian grammar framework.
\newblock In {\em 2012 International Conference on Asian Language Processing},
  93--96.
\newblock IEEE.

\bibitem[\protect\citeauthoryear{Gupta \bgroup et al\mbox.\egroup
  }{2016}]{gupta2016desire}
Gupta, A.; Akula, A.; Dasgupta, G.; Aggarwal, P.; and Mohapatra, P.
\newblock 2016.
\newblock Desire: Deep semantic understanding and retrieval for technical
  support services.
\newblock In {\em International Conference on Service-Oriented Computing},
  207--210.
\newblock Springer.

\bibitem[\protect\citeauthoryear{Kingma and Ba}{2015}]{kingma2014adam}
Kingma, D.~P., and Ba, J.
\newblock 2015.
\newblock Adam: A method for stochastic optimization.
\newblock {\em International Conference on Learning Representations (ICLR)}.

\bibitem[\protect\citeauthoryear{LeThanh, Abeysinghe, and
  Huyck}{2004}]{lethanh2004generating}
LeThanh, H.; Abeysinghe, G.; and Huyck, C.
\newblock 2004.
\newblock Generating discourse structures for written texts.
\newblock In {\em Proceedings of the 20th international conference on
  Computational Linguistics},  329.
\newblock Association for Computational Linguistics.

\bibitem[\protect\citeauthoryear{Mann and Thompson}{1988}]{mann1988rhetorical}
Mann, W.~C., and Thompson, S.~A.
\newblock 1988.
\newblock Rhetorical structure theory: Toward a functional theory of text
  organization.
\newblock {\em Text-Interdisciplinary Journal for the Study of Discourse}
  8(3):243--281.

\bibitem[\protect\citeauthoryear{Mikolov \bgroup et al\mbox.\egroup
  }{2013}]{mikolov2013distributed}
Mikolov, T.; Sutskever, I.; Chen, K.; Corrado, G.~S.; and Dean, J.
\newblock 2013.
\newblock Distributed representations of words and phrases and their
  compositionality.
\newblock In {\em Advances in neural information processing systems},
  3111--3119.

\bibitem[\protect\citeauthoryear{Palakurthi \bgroup et al\mbox.\egroup
  }{2015}]{palakurthi2015classification}
Palakurthi, A.; Ruthu, S.; Akula, A.; and Mamidi, R.
\newblock 2015.
\newblock Classification of attributes in a natural language query into
  different sql clauses.
\newblock In {\em Proceedings of the International Conference Recent Advances
  in Natural Language Processing},  497--506.

\bibitem[\protect\citeauthoryear{Pulijala, Akula, and
  Syed}{2013}]{pulijala2013web}
Pulijala, V.; Akula, A.~R.; and Syed, A.
\newblock 2013.
\newblock A web-based virtual laboratory for electromagnetic theory.
\newblock In {\em 2013 IEEE Fifth International Conference on Technology for
  Education (t4e 2013)},  13--18.
\newblock IEEE.

\bibitem[\protect\citeauthoryear{R~Akula \bgroup et al\mbox.\egroup
  }{2019}]{r2019natural}
R~Akula, A.; Todorovic, S.; Y~Chai, J.; and Zhu, S.-C.
\newblock 2019.
\newblock Natural language interaction with explainable ai models.
\newblock In {\em Proceedings of the IEEE Conference on Computer Vision and
  Pattern Recognition Workshops},  87--90.

\bibitem[\protect\citeauthoryear{Simonyan and
  Zisserman}{2014}]{simonyan2014very}
Simonyan, K., and Zisserman, A.
\newblock 2014.
\newblock Very deep convolutional networks for large-scale image recognition.
\newblock {\em arXiv preprint arXiv:1409.1556}.

\bibitem[\protect\citeauthoryear{Soricut and Marcu}{2003}]{soricut2003sentence}
Soricut, R., and Marcu, D.
\newblock 2003.
\newblock Sentence level discourse parsing using syntactic and lexical
  information.
\newblock In {\em Proceedings of the 2003 Conference of the North American
  Chapter of the Association for Computational Linguistics on Human Language
  Technology-Volume 1},  149--156.
\newblock Association for Computational Linguistics.

\bibitem[\protect\citeauthoryear{Thomason \bgroup et al\mbox.\egroup
  }{2014}]{thomason2014integrating}
Thomason, J.; Venugopalan, S.; Guadarrama, S.; Saenko, K.; and Mooney, R.~J.
\newblock 2014.
\newblock Integrating language and vision to generate natural language
  descriptions of videos in the wild.
\newblock In {\em Coling}, volume~2, ~9.

\bibitem[\protect\citeauthoryear{Yu \bgroup et al\mbox.\egroup
  }{2016}]{yu2016video}
Yu, H.; Wang, J.; Huang, Z.; Yang, Y.; and Xu, W.
\newblock 2016.
\newblock Video paragraph captioning using hierarchical recurrent neural
  networks.
\newblock In {\em Proceedings of the IEEE conference on computer vision and
  pattern recognition},  4584--4593.

\end{thebibliography}
\bibliographystyle{aaai}
\end{document}